%% file: root.tex
\pdfoutput=1
\documentclass[letterpaper, 10 pt, conference]{ieeeconf} 
\usepackage{amsmath} 
\usepackage{amsfonts,amssymb} 
\usepackage{comment} 
\usepackage{graphicx} 
\usepackage[labelformat=parens]{subcaption} 
\usepackage{float} 
\usepackage[inkscapelatex=false]{svg} 
\usepackage{array} 
\usepackage{makecell} 
\usepackage{caption} 
\usepackage{emf} 
\usepackage{threeparttable} 
\usepackage{booktabs} 
\usepackage{multirow} 
\usepackage{amssymb} 
\usepackage{verbatim} 
\usepackage{stfloats} 
\usepackage{cite} 
\makeatletter             
\let\NAT@parse\undefined  
\makeatother              
\usepackage{hyperref} 
\usepackage[acronym]{glossaries} 

\usepackage{afterpage}
\usepackage[commandnameprefix=always]{changes}

\IEEEoverridecommandlockouts                              
\title{\LARGE \bf
Residual Koopman Model Predictive Control for Enhanced Vehicle Dynamics with Small On-Track Data Input
}

\author{
Yonghao Fu\textsuperscript{*}, 
Cheng Hu\textsuperscript{*}, 
Haokun Xiong\textsuperscript{*}, 
Zhanpeng Bao, 
Wenyuan Du,
Edoardo Ghignone,\\
Michele Magno,
Lei Xie\textsuperscript{†}, 
and Hongye Su
\thanks{This work was supported by the Ningbo Key Research and Development Plan (No.2023Z116).}
\thanks{\textsuperscript{*}These authors contributed equally to this work.}
\thanks{\textsuperscript{†}The corresponding author of this paper.}
\thanks{Yonghao Fu, Cheng Hu, Haokun Xiong, Zhanpeng Bao, Wenyuan Du, Lei Xie, and Hongye Su are with the State Key Laboratory of Industrial, Zhejiang University, Hangzhou 310027, China. Emails: \{22360414, 22032081, 12332029\}@zju.edu.cn; zhanpbao@163.com; wenyuandu@zju.edu.cn; \{leix, hysu\}@iipc.zju.edu.cn.}
\thanks{Edoardo Ghignone, and Michele Magno are associated with ETH Zürich. Emails: \{edoardo.ghignone, michele.magno\}@pbl.ee.ethz.ch.}%
}

\let\oldtwocolumn\twocolumn
\renewcommand\twocolumn[1][]{%
    \oldtwocolumn[{#1}{
    \begin{center}
           \includegraphics[width=\textwidth]{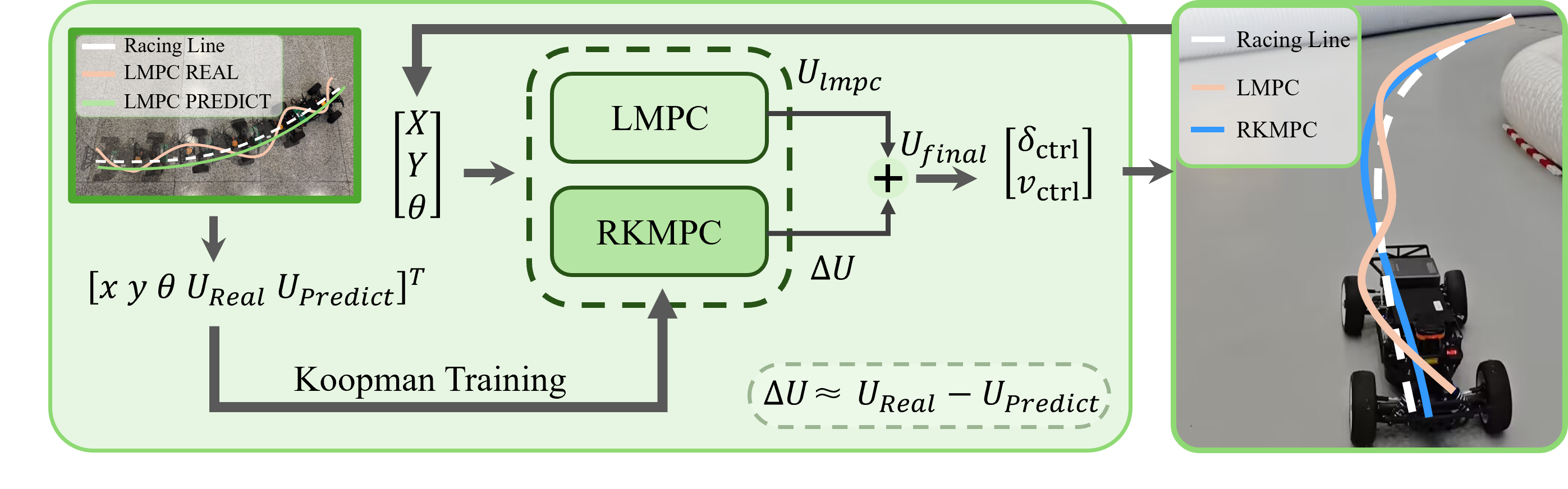}
           \captionof{figure}{The proposed RKMPC framework uses two linear MPC to calculate control inputs: a linear MPC computes the baseline control $U_{lmpc}$ based on the vehicle kinematics model, and a neural network-based residual Koopman MPC computes the compensation $\Delta U$. The final control command is obtained by adding these two components. Compared to traditional Koopman-MPC, this approach reduces training data requirements and enhances control performance.}
           \label{fig.Header image}
        \end{center}
    }]
}

\begin{document}

\include{acr} 

\maketitle

\thispagestyle{empty}
\pagestyle{empty}


\begin{abstract}

In vehicle trajectory tracking tasks, the simplest approach is the \gls{pp} Control. However, this single-point preview tracking strategy fails to consider vehicle model constraints, compromising driving safety. \gls{mpc} as a widely adopted control method, optimizes control actions by incorporating mechanistic models and physical constraints. While its control performance critically depends on the accuracy of vehicle modeling.
Traditional vehicle modeling approaches face inherent trade-offs between capturing nonlinear dynamics and maintaining computational efficiency, often resulting in reduced control performance.
To address these challenges, this paper proposes 
\gls{rkmpc} framework. This method uses two linear \gls{mpc} architecture to calculate control inputs: a \gls{lmpc} computes the baseline control input based on the vehicle kinematic model, and a neural network-based \gls{rkmpc} calculates the compensation input.
The final control command is obtained by adding these two components. 
This design preserves the reliability and interpretability of traditional mechanistic model while achieving performance optimization through residual modeling.
This method has been validated on the Carsim-Matlab joint simulation platform and a physical 1:10 scale F1TENTH racing car.
Experimental results show that \gls{rkmpc} requires only 20\% of the training data needed by traditional \gls{kmpc} while delivering superior tracking performance. 
Compared to traditional \gls{lmpc}, \gls{rkmpc} reduces lateral error by 11.7\%–22.1\%, decreases heading error by 8.9\%–15.8\%, and improves front-wheel steering stability by up to 27.6\%. The implementation code is available at: \href{https://github.com/ZJU-DDRX/Residual_Koopman}{https://github.com/ZJU-DDRX/Residual\_Koopman}.

\end{abstract}



\section{INTRODUCTION}


In vehicle trajectory tracking tasks, \gls{pp} Controller is a commonly used model-free algorithm. It is easy to implement, but also has some limitations. On the one hand, it only relies on single-point preview, lacking the ability to globally plan for the future trajectory. On the other hand, failing to consider the vehicle's mechanism model constraint may cause the control output to exceed the physically feasible range, thereby threatening driving safety.
In contrast, \gls{mpc} significantly enhances the safety and robustness of control by establishing a mechanism model of the vehicle and incorporating physical constraints. Its core idea is to use a rolling optimization strategy to solve the optimal control sequence within a finite time horizon at each time step.
However, \gls{mpc} performance is fundamentally limited by the inherent difficulty of accurately modeling nonlinear vehicle dynamics. Complex factors like tire mechanics and suspension characteristics are hard to characterize precisely, often causing discrepancies between predicted and actual responses\cite{8315037}.

Current modeling approaches primarily utilize kinematic or dynamic models\cite{kong2015kinematic}: kinematic models efficiently describe geometric motion relationships and are preferred for trajectory tracking applications, while dynamic models offer higher precision but demonstrate greater parameter sensitivity and computational complexity.

These traditional methods face inherent challenges - as strongly nonlinear systems, vehicle state evolution depends on multiple coupled physical factors. Traditional linearization techniques, while computationally convenient, inevitably introduce modeling errors\cite{kong2015kinematic}. The Koopman operator theory presents an innovative alternative by constructing data-driven linear representations in high-dimensional space that preserve nonlinear characteristics, particularly suitable for integration with \gls{mpc} frameworks\cite{korda2018linear}.

However, Koopman approaches exhibit two significant limitations. First, their requirement for extensive high-quality training data leads to prohibitively expensive implementation costs, particularly in racing scenarios\cite{cibulka2020model}. Second, the purely data-driven framework lacks integration with vehicle mechanism knowledge, compromising both model reliability and safety assurance\cite{sinha2020robust}.

To overcome these limitations, we propose the \gls{rkmpc} framework that establishes a residual model using neural network-based Koopman operators, with the vehicle kinematic model serving as the baseline framework. Our approach employs data-driven techniques to learn residual control inputs, achieving an optimal balance between mechanistic modeling and data efficiency. 
The key innovations of this work include:

\begin{itemize}

\item \textbf{A novel residual Koopman framework}. \gls{rkmpc} framework uses two linear \gls{mpc} architecture to calculate control inputs: a linear \gls{mpc} computes the baseline control based on the vehicle kinematics model, and a neural network-based residual Koopman \gls{mpc} computes the compensation. The final control command is obtained by adding these two components. Compared to traditional \gls{kmpc}, this approach reduces training data requirements and enhances control performance. 
An overview of the method is given in Fig. \ref{fig.Header image}.


\item \textbf{Comprehensive and reliable Simulation and Real-vehicle experiments}. 
Under small-data conditions our \gls{rkmpc} framework shows better performance through its residual koopman framework. Validation on both Carsim-Matlab co-simulation and a 1:10 scale physical vehicle shows that \gls{rkmpc} achieves 11.7\%-22.1\% lateral error reduction, 8.9\%-15.8\% heading error decrease, and up to 27.6\% steering stability improvement compared to traditional \gls{lmpc}, while achieving comparable performance with only 20\% of the training data required by traditional \gls{kmpc} approaches.

\item \textbf{Open-source onboard algorithm}. 
The proposed algorithm is applied on the official F1TENTH hardware platform and is fully open-source on GitHub.

\end{itemize}

\section{RELATED WORKS}

\subsection{Traditional Vehicle Trajectory Tracking Methods}

In vehicle trajectory tracking, \gls{pp} control is a commonly used model-free algorithm. However, this approach cannot effectively handle vehicle model constraints and keep driving safety \cite{hoffmann2007autonomous,wang2017improved}.


To address this issue, \gls{mpc} has gradually become the mainstream strategy for vehicle trajectory tracking\cite{li2024data,hu2022combined}. 
\gls{mpc} optimizes the control inputs over a prediction horizon in each control cycle, minimizing trajectory tracking errors while satisfying vehicle model constraints. \gls{mpc} includes both linear and nonlinear variants. \gls{lmpc} reduces the computation time by linearizing the model, but this process inevitably sacrifices some model accuracy. \gls{nmpc} directly solves nonlinear problems without linearization, but demands high computational resources, especially for onboard processing \cite{bienemann2023model,gao2014tube}. Residual control enhances traditional methods with learning-based components. Zhang et al. \cite{zhang2022residual} achieved efficient autonomous racing using residual policy learning with onboard sensors. Trumpp et al. \cite{trumpp2023residual} improved controll
er adaptability and lap times through residual learning. Long et al. \cite{long2025physical} combined neural networks with physics models to improve prediction accuracy and data efficiency.


\subsection{Koopman-Based Model Predictive Control}

With the development of the Koopman operator theory in recent years, \gls{kmpc} has gained attention. The Koopman operator is a tool that maps nonlinear systems to a high-dimensional space and constructs linear models to describe the nonlinear evolution of the system\cite{brunton2016koopman}. Through this approach, \gls{kmpc} maintains high computational efficiency while not sacrificing system accuracy. Compared to traditional \gls{lmpc} and \gls{nmpc}, \gls{kmpc} can handle complex nonlinear systems without additional linearization, effectively addressing the high computational complexity of \gls{nmpc}. It provides accuracy comparable to \gls{nmpc} while maintaining a lower computational burden. In several simulation and real-vehicle experiments, \gls{kmpc} has demonstrated good control performance, maintaining high precision in complex dynamic environments
\cite{cibulka2020model,xiao2022deep}.

Despite the significant advantages demonstrated in theory and practice, some drawbacks remain. First, the method requires a large amount of training data, and collecting such data is particularly costly in racing scenarios compared to normal road vehicles\cite{cibulka2020model}. Second, as a purely data-driven approach without incorporating physical models, it cannot ensure system safety and stability\cite{sinha2020robust}.

To overcome these limitations, this paper proposes 
\gls{rkmpc} framework. This design preserves the reliability and interpretability of traditional mechanistic model while achieving performance optimization through data-driven koopman residual modeling.


\subsection{Comparison of Different Control Methods}
As a summary, we list the characteristics of different model-based control methods in Table \ref{tab:control_methods}. \textbf{Model} indicates whether the control approach adopts a mechanistic or data-driven model. \textbf{Compute} indicates the approximate computing time required. \textbf{Data} indicates whether offline data needs to be collected for training.

\begin{table}[htbp]
\caption{Comparison of control methods in terms of model dependency, computation time, and data requirements.}
\label{tab:control_methods}
\scriptsize
\renewcommand{\arraystretch}{1.5}
\centering
\begin{threeparttable}
\begin{tabular}{@{}l|c|c|l@{}}
\toprule
\textbf{Method} & \textbf{Model} & \textbf{Compute (ms)\textsuperscript{*}} & \textbf{Data\textsuperscript{†}}  \\
\midrule
\textbf{\gls{lmpc}}\cite{yakub2015comparative} & Mechanism & 1--15 & None  \\
\textbf{\gls{nmpc}}\cite{gao2014tube} & Mechanism & 10--70 & None  \\
\textbf{\gls{kmpc}}\cite{korda2018linear} & Data Driven & 1--20 & Rich  \\
\textbf{\gls{rkmpc} (Ours)} & Mechanism + Data Driven & 1--20 & Small  \\
\bottomrule
\end{tabular}
\begin{tablenotes}
\footnotesize
\item[*] The computation time is based on simulations running on a Windows system with an i5-13500HX CPU.
\item[†]\textbf{None} means no dataset is required, \textbf{Rich} requires a large dataset, and \textbf{Small} needs only a few laps of on-track data. In our experiment, \gls{rkmpc} uses approximately 8,000 data points, while \gls{kmpc} used about 50,000 data points.
\end{tablenotes}
\end{threeparttable}
\end{table}

This paper consists of several sections. Section III introduces traditional \gls{lmpc} and the Koopman method. Section IV explains the \gls{rkmpc} method, including data preprocessing and the \gls{rkmpc} control structure. Sections V and VI present the application of \gls{rkmpc} in simulation and on the F1TENTH vehicle. Section VII serves as a conclusion that summarizes the findings and outlines directions for future research.

\section{PRELIMINARIES}

This section will introduce the method for obtaining the nominal vehicle kinematic model and the Koopman model and provide an example to explain how the Koopman \gls{edmd} algorithm is implemented in nonlinear systems. Subsequently, we will combine the nominal and Koopman models to form the residual Koopman models.
\subsection{Nominal vehicle model}


In this paper, we adopt the kinematic bicycle vehicle model. Compared with dynamic models, this model requires fewer parameters and is applicable to most scenarios\cite{wu2024learning}. The specific equations are as follows:
\begin{equation}
    \label{kinematic function}
    \begin{aligned} 
        \dot{x} & =v \cos (\theta) \\
        \dot{y} & =v \sin (\theta) \\
        \dot{\theta} & =\frac{v}{L} \tan (\delta)
    \end{aligned}
\end{equation}
where $x$ and $y$ are the global positions, $\theta$ is the yaw angle, $\delta$ is the steering angle, $v$ is the velocity, $L$ is the wheelbase,the symbol [ \textbf{$\dot{}$} ]  represents the rate of change of that variable with respect to time. .

The sampling time can be set to $T$ to discretize and linearize Eq. (\ref{kinematic function}). The resulting discrete linear model are Eq. (\ref{Linear MPC function1}) and Eq. (\ref{Linear MPC function2}):
\begin{equation}
    \label{Linear MPC function1}
    \overline{\boldsymbol{\xi}}_{\mathrm{kin}}(k+1)=\boldsymbol{A}_{\mathrm{kin}}(k) \overline{\boldsymbol{\xi}}_{\mathrm{kin}}(k)+\boldsymbol{B}_{\mathrm{kin}}(k) \tilde{\boldsymbol{u}}_{\mathrm{kin}}(k)
\end{equation}
\begin{equation}
    \label{Linear MPC function2}
        \begin{gathered}    \overline{\boldsymbol{\xi}}_{\mathrm{kin}}=\left[\begin{array}{c}x-x_{\mathrm{r}} \\ y-y_{\mathrm{r}} \\ \theta-\theta_{\mathrm{r}}\end{array}\right] \\ \boldsymbol{A}_{\mathrm{kin}}(k)=\left[\begin{array}{ccc}1 & 0 & -v_{\mathrm{r}} \sin \theta_{\mathrm{r}} T \\ 0 & 1 & v_{\mathrm{r}} \cos \theta_{\mathrm{r}} T \\ 0 & 0 & 1\end{array}\right], \\ \boldsymbol{B}_{\mathrm{kin}}(k)=\left[\begin{array}{cc}\cos \theta_{\mathrm{r}} T & 0 \\ \sin \theta_{\mathrm{r}} T & 0 \\ \frac{\tan \delta_{\mathrm{f}, \mathrm{r}} T}{l} & \frac{v_{\mathrm{r}} T}{l \cos ^2\left(\delta_{\mathrm{f}, \mathrm{r}}\right)}\end{array}\right],
    \end{gathered}
\end{equation}

where $\overline{\boldsymbol{\xi}}_{\mathrm{kin}}$ is the state vector, representing the deviations of $x$, $y$, and $\theta$ from the reference trajectory, $\tilde{\boldsymbol{u}}_{\mathrm{kin}}$ is the input vector, including the control variables $\delta$ and $v$, $\mathbf{A}_{\text{kin}}(k)$ and $\mathbf{B}_{\text{kin}}(k)$ are the state transition and input matrices respectively, $k$ is the k-th time step, the subscript $r$ denotes the reference values.

\subsection{Approximating the Data-driven Koopman Operator}

Assume a discrete-time nonlinear dynamic system with the state update equation \( x^+ = f(x, u) \), where \( x \) represents the state variables, \( u \) represents the control inputs, and \( f(\cdot) \) is the nonlinear state equation. To address the nonlinearities in the system, we utilize the Koopman operator, which linearizes the nonlinear dynamics. Specifically, by using a set of observation functions \( g(x_t) \), the system's state is lifted from a low-dimensional space to a high-dimensional space\cite{korda2018linear}. we define the Koopman operator $\mathcal{K} g$ in the following form, which acts on the nonlinear state update equation $f(x_t, u_t)$ through the observation functions $g(x_t)$ to achieve state space lifting and representation, as shown in Eq. (\ref{1}):
\begin{equation}
\label{1}
    \mathcal{K} g\left(x_t\right)=g\circ f(x_t,u_t)=g\left(f\left(x_t, u_t\right)\right) 
\end{equation}
 
where $\circ$ is the composition operator, and $g(x_t)$ represents lifting the state variables of the system from the original $n$-dimensional space $\mathbb{R}^n$ to a higher $m$-dimensional space $\mathbb{R}^m$.

Since the Koopman operator is infinite-dimensional, but this is not feasible in practical applications, a finite number of observation functions are used for approximation\cite{mauroy2020introduction}. By recording the system's state and control input sequences, appropriate observation functions are employed to lift the system's state to a higher-dimensional space, obtaining the high-dimensional state-space representation \( z_t \), which provides an approximation of the Koopman operator.

To solve this, the optimization problem is formulated to minimize the state transition error. The optimal matrices \( A \) and \( B \) are then computed using least squares, representing the linearized system, as shown in Eq. (\ref{2}):
\begin{equation}
\label{2}
\min _{A, B} \sum_{t=1}^K\left\|z_{t+1}-\left(A  z_t+B  u_t\right)\right\|_2^2
\end{equation}

To further map the high-dimensional state back to the original state, we introduce a matrix \( C \), establishing the relationship between the high-dimensional space and the original state space, as shown in Eq. (\ref{3}):
\begin{equation}
\label{3}
\min _C \sum_{t=1}^K\left\|x_t-C  z_t\right\|_2^2
\end{equation}
These matrices can be solved using the pseudo-inverse method described in \cite{korda2018linear}.
\begin{equation} 
\label{A more ideal method for solving A and B} 
[A, B]=z_{t+1}\left[\begin{array}{l}z_t \ U\end{array}\right]\left(\left[\begin{array}{l}z_t \ U\end{array}\right]^T\left[\begin{array}{l}z_t \ U\end{array}\right]\right)^{\dagger} 
\end{equation}
Finally, the high-dimensional state-space equation is obtained as shown in Eq. (\ref{Koopman linear state equation}).
\begin{equation}
    \label{Koopman linear state equation}
    \begin{aligned} 
        z_{t+1} & =A z_t+B u_t \\
        x_t & =C z_t
    \end{aligned}
\end{equation}

\section{RESIDUAL KOOPMAN CONTROL METHOD}

This section will provide a detailed description of the structure of the Data Preprocessing Process and the Control Structure of \gls{rkmpc}.

\subsection{Data Collection and Preprocessing}

\begin{figure*}[htbp]
  \centering
  \includegraphics[width = 14cm]{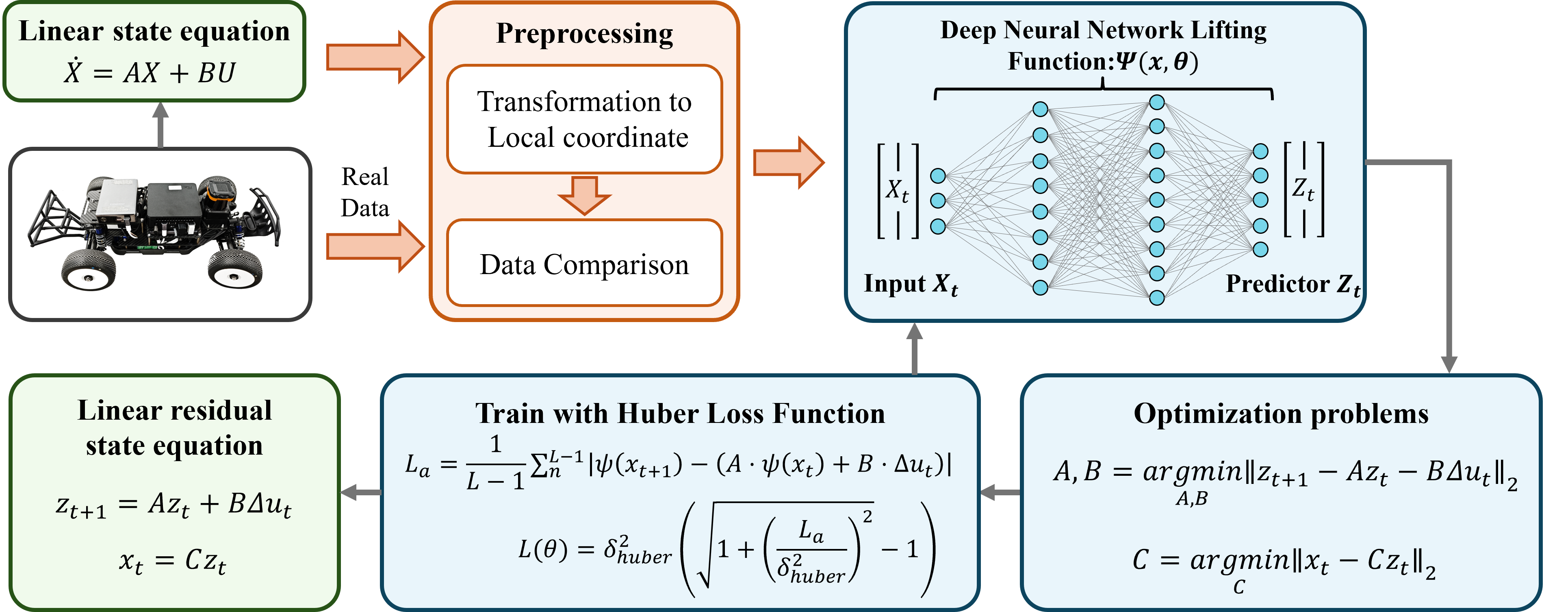}
  \caption{Framework for data collection and preprocessing: combining local coordinate transformation, neural network training, and optimization for linear residual state quation.}
  \label{fig.Data Collection and Preprocessing}
  \vspace{-1em}
\end{figure*}

The typical Koopman method requires a lot of actual vehicle data for training to obtain a relatively accurate model. However, challenges such as insufficient existing data and high data collection costs often exist for racing vehicles or special-purpose vehicles. Therefore, we adopt a method combined with a mechanism-based model, focusing on residual on-track data. 
\begin{figure}[H]
  \centering
  \includegraphics[width = 7cm]{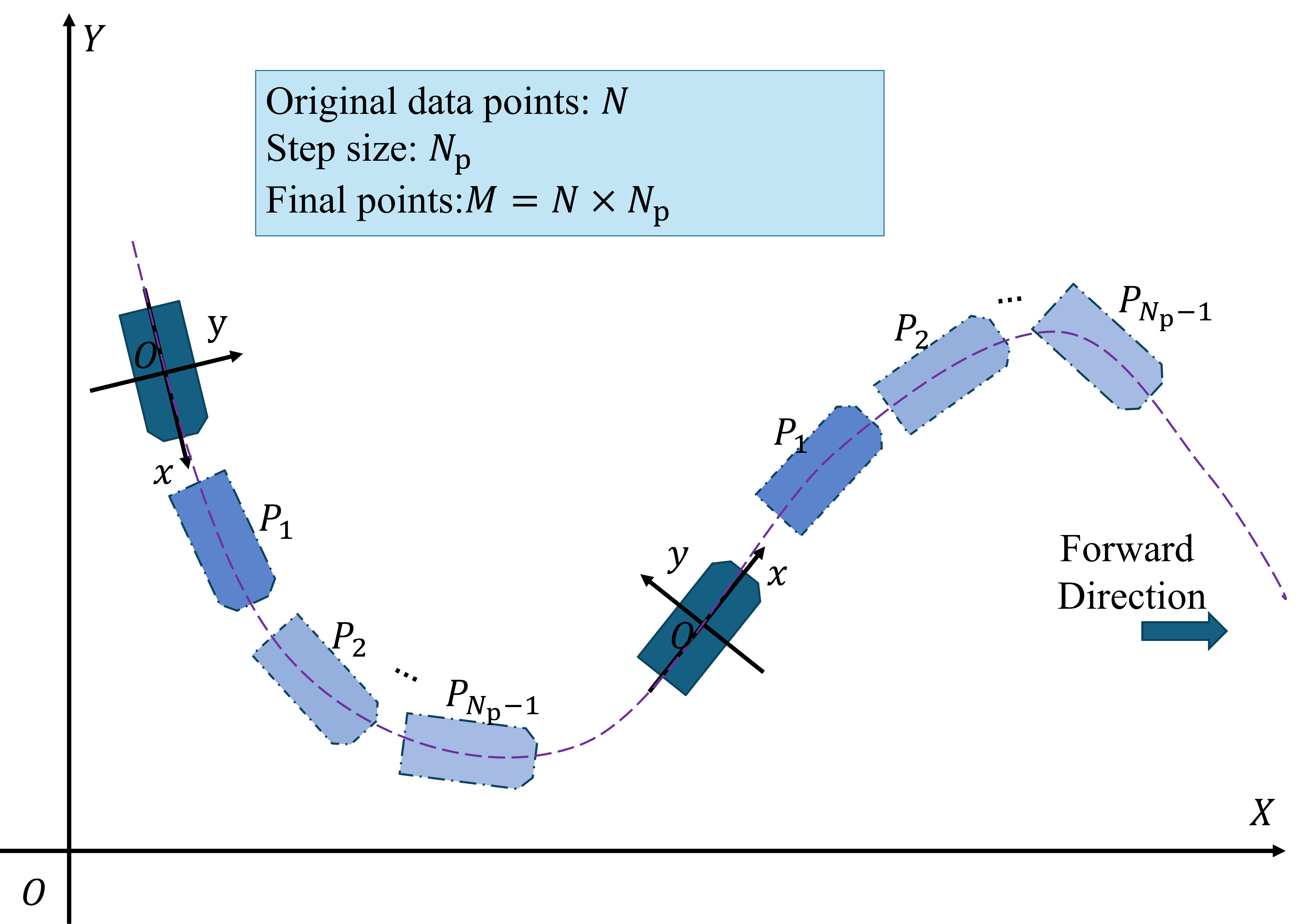}
  \caption{Local coordinate transformation ensures consistent linearization by aligning the vehicle's heading angle to zero and compressing the data range for efficient Koopman control training.}
  \label{fig.Coordinate Transformation}
\end{figure}
This approach, outlined in Fig. \ref{fig.Data Collection and Preprocessing}, can significantly reduce the required data volume and training time.

After collecting continuous-time data, a certain number of random points need to be selected for the data transformation to a local coordinate system. This approach has two advantages: first, it ensures that during the linearization process of the vehicle kinematic equations, the reference heading angle at the operating point is always zero, making the form of the state equations at each position as consistent as possible; second, it compresses the data range, eliminating the need for normalization later.

The specific operation is shown in Fig. \ref{fig.Coordinate Transformation}. The raw data collected is based on the global coordinate system. We select a proportion of points as the coordinate origin dataset based on a proportion, which we call the conversion ratio, and select \( N_{p}-1 \) points (\( P_1, P_2, \dots, P_{N_{p}-1} \)) following the time sequence, then perform a coordinate transformation on each of them. If the total number of original data points is \( N \), then after the coordinate transformation, the total number of data points will be \(M = N \times N_p\), where \( N_p \) is the number of consecutive points selected, it should be slightly greater than the \gls{mpc} prediction horizon, and the proportion of the coordinate origin data set is related to the original data's size.
\begin{figure}[htbp]
  \centering
  \includegraphics[width = 6cm]{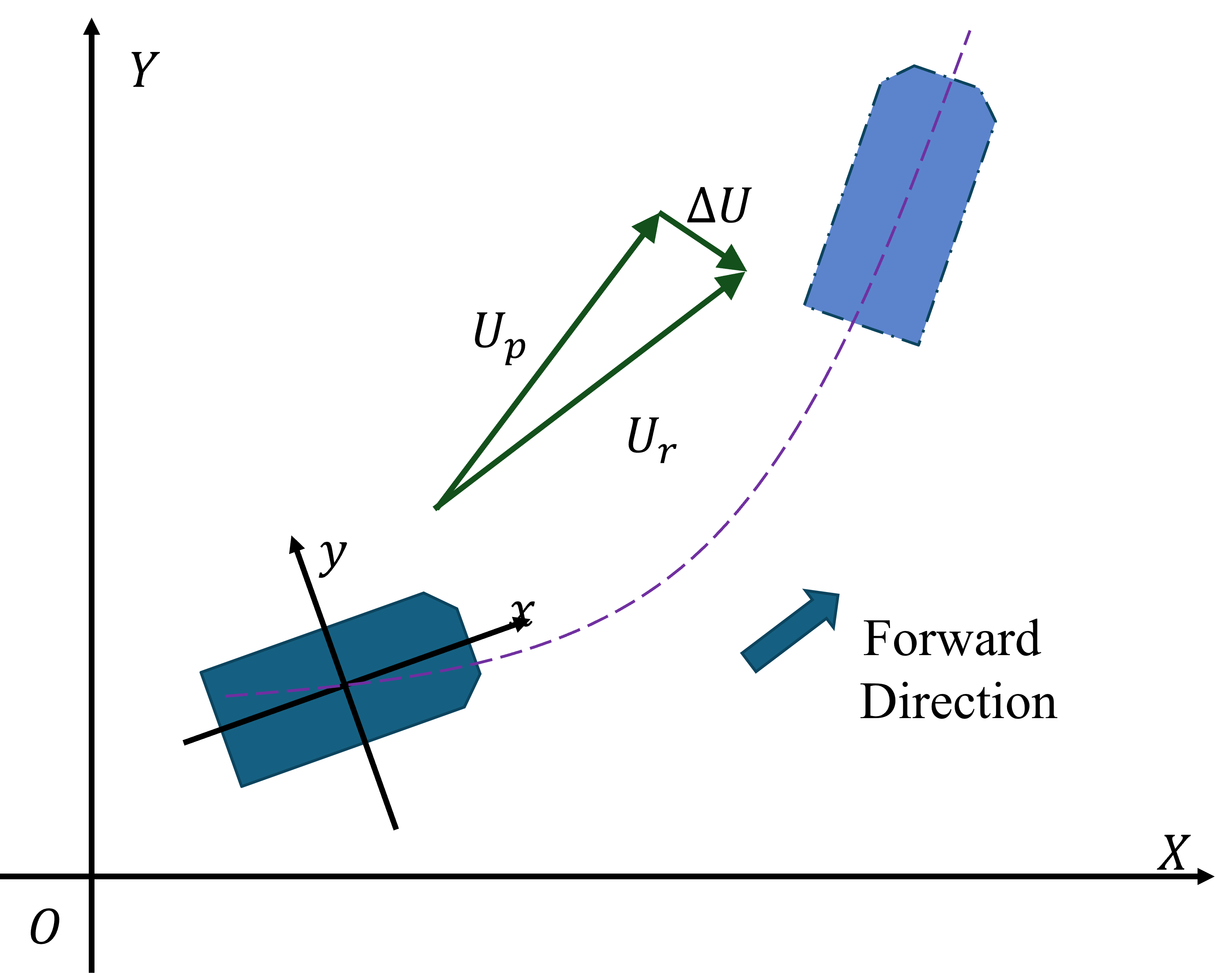}
  \caption{Representation of control residual \( \Delta U = U_r - U_p \) for correcting predicted control inputs.}
  \label{fig.Control residual form}
\end{figure}

After obtaining the data from the coordinate transformation,  the next step is to convert the control inputs into control residuals, as shown in Fig. \ref{fig.Control residual form}. For the same state evolution, the control input predicted by the \gls{mpc} is \( U_p \), but the control input that should be executed is \( U_r \) due to the linearization errors. Therefore, a control residual \( \Delta U = U_r - U_p \) is needed as the correction. After processing each data point in this manner, the Koopman data in the format of \( [X, \dot{X}, \Delta U] \) can be obtained. In this paper, the steering angle and velocity are adopted as control inputs, that is $u=[v,\delta]$.
\subsection{Construction of the Control Structure}
\begin{figure*}[htbp]
  \centering
  \includegraphics[width = 14cm]{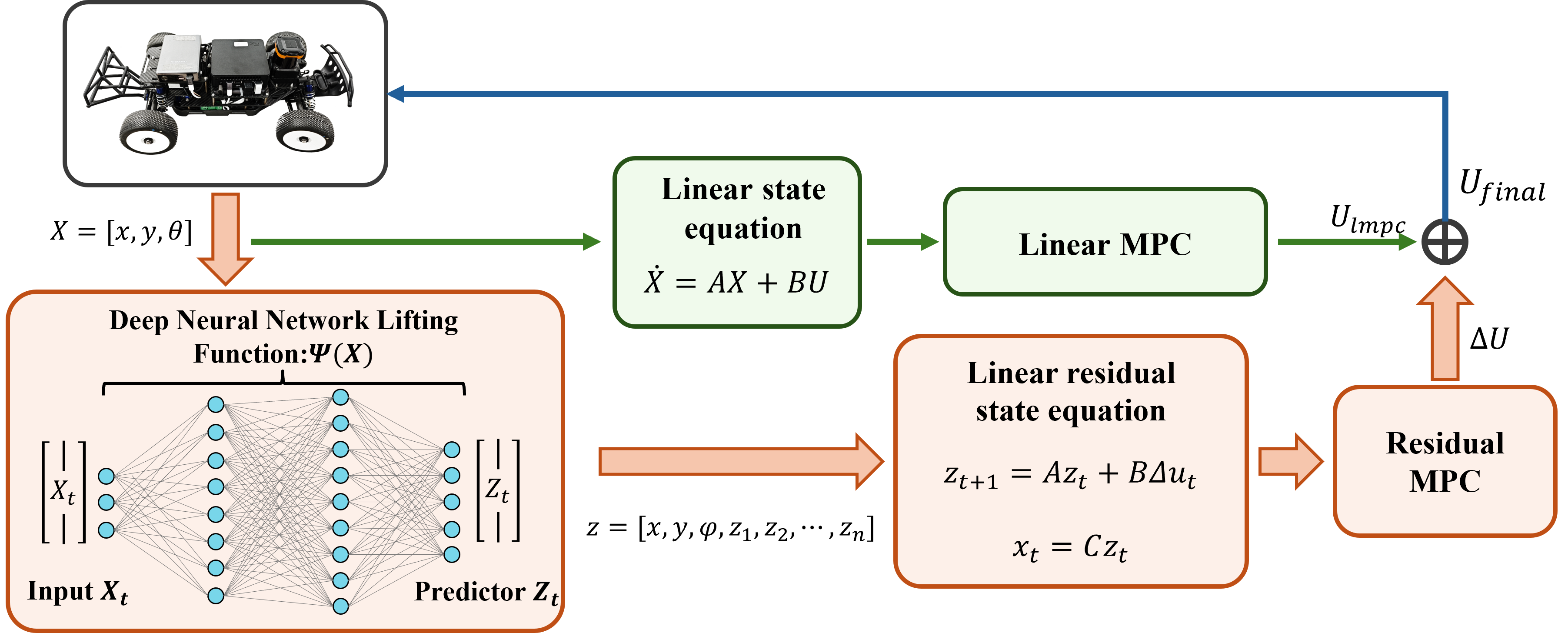}
  \caption{\gls{rkmpc} structure: integration of neural network lifted state representations with \gls{lmpc} and \gls{rkmpc}}
  \label{fig.Construction of the Control Structure}
\end{figure*}

In this paper, we will use a high-dimensional neural network as the Koopman operator's basis function to capture the residual system's nonlinear behavior and accurately approximate the nonlinear dynamics as a linear system over the lifted state space. The neural network takes the vehicle state $(x, y, \theta)$ as input and consists of 2 fully connected layers, each using ReLU activation. After obtaining the appropriate corresponding \(A\) and \(B\) matrices through Eq. (\ref{A more ideal method for solving A and B}), we also need to optimize the basis functions to reduce the loss. Therefore, the loss function of the neural network can be set as in Eq. (\ref{net loss function}). To enhance the robustness of the training process, we incorporate the Huber loss function as the final cost value, where \(\delta_{huber}\) is the hyperparameter of the loss function, controlling the trade-off between Mean Squared Error (MSE) and Mean Absolute Error (MAE)\cite{wang2021deep}. When the cost value is sufficiently small, the state equation can be considered consistent with the true state of the system and exhibits linearity.
\begin{equation}
    \label{net loss function}
    \begin{gathered}
        L_a=\frac{1}{L-1} \sum_n^{L-1}\left|\psi\left(x_{t+1}\right)-\left(A \psi\left(x_t\right)+B \Delta u_t\right)\right| \\ L(\theta)=\delta_{huber}^2\left(\sqrt{1+\left(\frac{L_a}{\delta_{huber}}\right)^2}-1\right)
    \end{gathered}
\end{equation}

Finally, after continuous optimization, we can obtain the state equation in the following form:
\begin{equation}
    \begin{aligned} 
        z_{t+1} & =A z_t+B \Delta u_t \\
        x_t & =C z_t
    \label{eq.recidual state euqation}
    \end{aligned}
\end{equation}
where $z = [x, y, \theta, z_1, z_2, \dots, z_n]$ represents the high-dimensional state variable obtained by applying the lift function.

A scheme summarizing the \gls{rkmpc} structure is available in Fig. \ref{fig.Construction of the Control Structure}.

After obtaining the residual state equation, it can be integrated into the existing \gls{lmpc} controller. After acquiring the vehicle's state variables, on the one hand, the \gls{lmpc} control is directly applied to obtain the original control input \(U_0\). On the other hand, the state variables are passed through a \gls{dnn} for dimensionality lifting, converting them into high-dimensional state variables. A residual \gls{mpc} controller is then constructed to obtain the compensatory control input \(\Delta U\). Finally, the two control inputs are added to obtain the total control input \(U_{final}\), the final control signal.
\begin{figure}[htbp]
  \centering
  \includegraphics[width = 7cm]{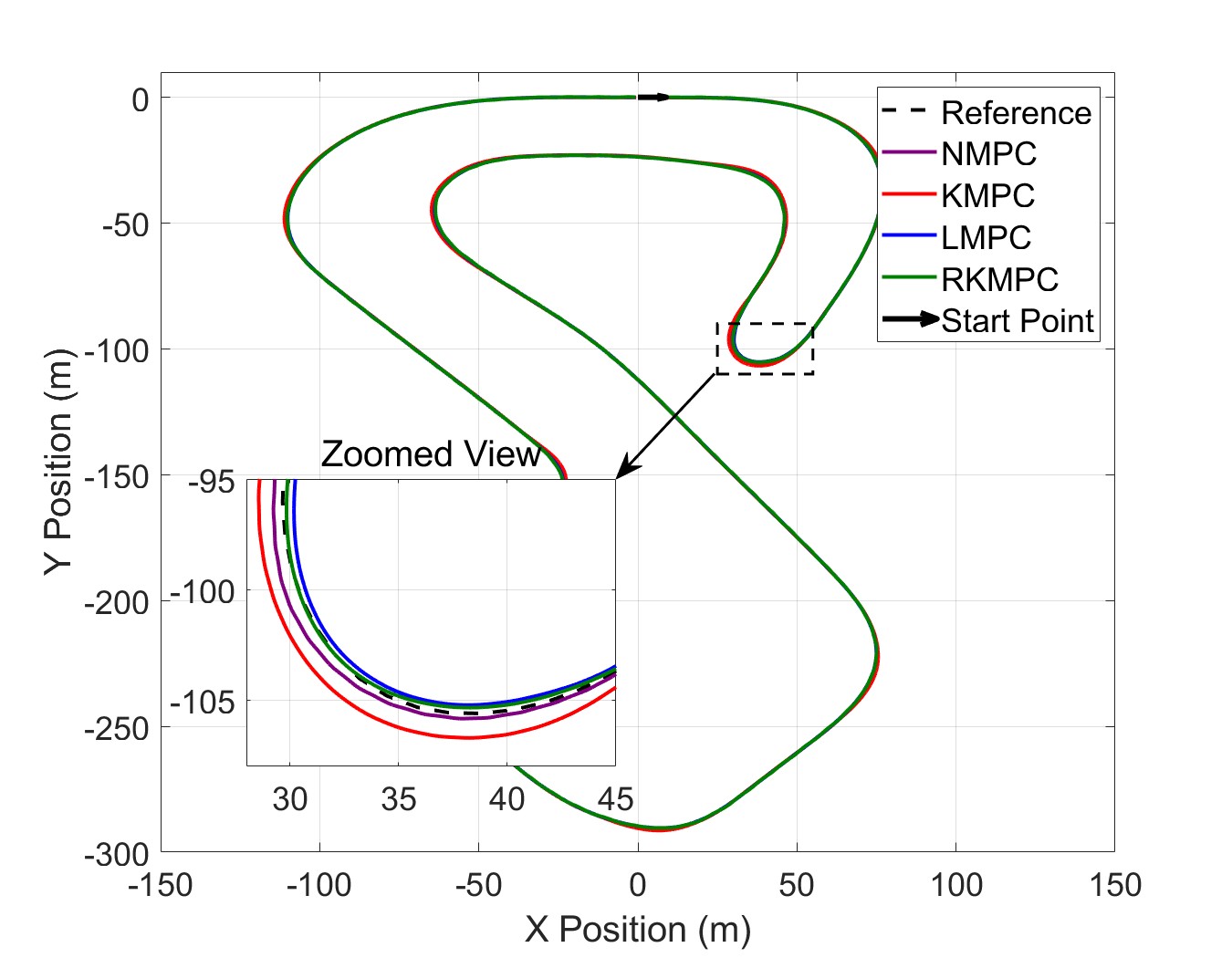}  
  \caption{Comparison of paths with different control methods in simulation.}
  \label{fig.racetrack-database}
\end{figure}
\begin{figure}[htbp]
  \centering
  \includegraphics[width = 7cm]{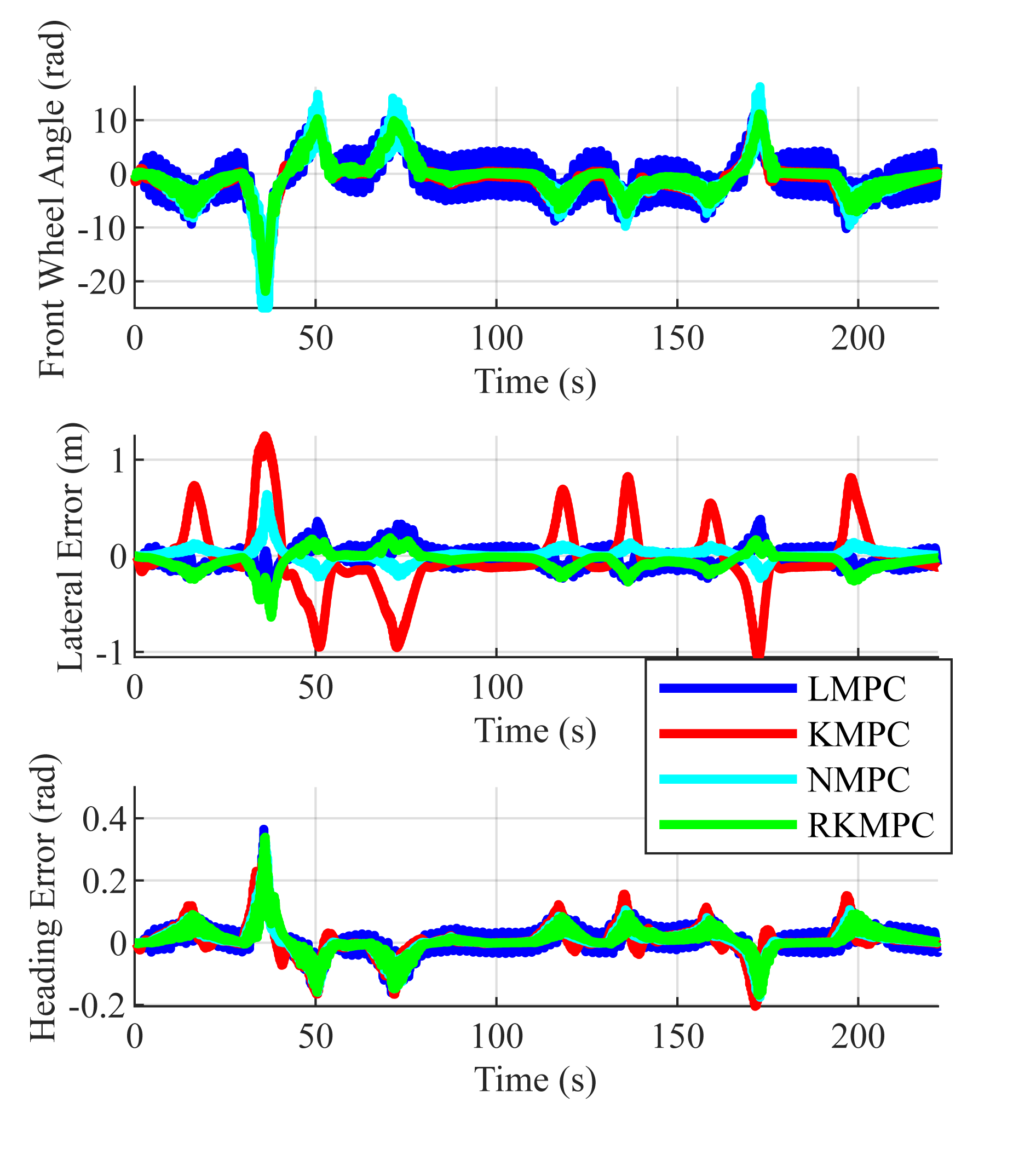}
  \caption{Comparison of front wheel angle, lateral error, and heading error among \gls{lmpc}, \gls{kmpc}, \gls{nmpc}, and \gls{rkmpc} in simulation, although \gls{kmpc} operates normally, its performance is hindered by the lack of data richness, preventing optimal results.}
  \label{fig.Comparison of the Three Controllers for One Lap}
\end{figure}

The \gls{lmpc} and \gls{rkmpc} parts need to be designed separately in the \gls{mpc} design process. For the linearized kinematic vehicle model, we adopt the linearized equations as described in Eq. (\ref{Linear MPC function2}). The final optimization problem is given in Eq. (\ref{MPC1}).
\begin{align}
    \min  \sum_{k=0}^{N-1} &\Big(
    \left(x_k-x_{\mathrm{ref}, k}\right)^2+\left(y_k-y_{\mathrm{ref}, k}\right)^2 \label{MPC1}\\
    +&\lambda \left(\theta_k-\theta_{\mathrm{ref}, k}\right)^2 
    +\mu\left(v_k-v_{\mathrm{ref}, k}\right)^2 \nonumber\\
    +&\epsilon\left(\delta-\delta_{\mathrm{ref}, k}  \right)^2 \Big)\nonumber\\
    \text{s.t. } x_{\min }& \leq x_k \leq x_{\max },\; y_{\min } \leq y_k \leq y_{\max } \nonumber\\
    \theta_{\min }& \leq \theta_k \leq \theta_{\max },\; v_{\min } \leq  v_k \leq  v_{\max } \nonumber\\
     \delta_{\min }& \leq  \delta_k \leq  \delta_{\max } \nonumber,\;\text{Vehicle Kinematic Model}(\ref{Linear MPC function1})
\end{align}
For the \gls{rkmpc} design, the control input is $\Delta \textbf{u}=[\Delta v, \Delta \delta]$. The final optimization problem is given in Eq. (\ref{MPC2}).
\begin{align}
    \min \sum_{k=0}^{N-1}&\Big(    \left(x_k-x_{\mathrm{ref}, k}\right)^2+\left(y_k-y_{\mathrm{ref}, k}\right)^2 \label{MPC2}\\
    +&\lambda \left(\theta_k-\theta_{\mathrm{ref}, k}\right)^2+\mu\left(\Delta v_k\right)^2+\epsilon\left(\Delta \delta_k\right)^2 \Big) \nonumber\\
    \text{s.t. }  x_{\min } &\leq x_k \leq x_{\max }, \; y_{\min } \leq y_k \leq y_{\max } \nonumber\\
    \theta_{\min } &\leq \theta_k \leq \theta_{\max }, \; \Delta v_{\min } \leq \Delta v_k \leq \Delta v_{\max } \nonumber\\
    \Delta \delta_{\min } &\leq \Delta \delta_k \leq \Delta \delta_{\max } \nonumber,\;\text{Recidual Koopman Model}(\ref{eq.recidual state euqation})
\end{align}

\section{SIMULATION RESULT}


In this section, we compare the performance of \gls{rkmpc}, \gls{kmpc}, \gls{lmpc}, and \gls{nmpc} in trajectory tracking tasks, as well as the data requirements of \gls{rkmpc} and \gls{kmpc}. These algorithms are verified using the Matlab-Carsim platform.
The map data used in the experiment is sourced from actual F1 race tracks and scaled according to a specific ratio\cite{racetrack-database}.


\begin{table*}[htbp]
\renewcommand{\arraystretch}{1.5}
\caption{Comparison of front wheel angle, lateral error, heading error and compuation time among \gls{lmpc}, \gls{kmpc}, \gls{nmpc}, and \gls{rkmpc} in simulation.}
\label{table_1}
\centering
\begin{tabular}{@{}l|c|c|c|c@{}}
\toprule
\textbf{Method} & \textbf{Lateral Error (m)} & \textbf{Heading Error (rad)} & \textbf{Front Wheel Angle Rate (rad/s)} & \textbf{Computation Time (ms)} \\ \midrule
\gls{lmpc} (baseline) & 0.1115               & 0.0475                  & 0.1570                  & $t_{\text{mean}} = 2.38$,  $t_{\text{max}} = 13.30$       \\ 
\gls{nmpc}            & \textbf{0.0808}               & 0.0443                  & 0.2414                  &$t_{\text{mean}} = 13.68$,  $t_{\text{max}} = 66.38$\\ 
\gls{kmpc}            & 0.1650               & 0.0436                  & 0.1794                  &$t_{\text{mean}} = 3.78$,  $t_{\text{max}} = 17.58$\\ 
\gls{rkmpc}           & 0.0990 ($\downarrow$ 11.21\%)      & \textbf{0.0434 ($\downarrow$ 8.63\%)}         & \textbf{0.1175 ($\downarrow$ 27.58\%)}         &$t_{\text{mean}} = 6.73$,  $t_{\text{max}} = 16.14$\\ 
\bottomrule
\end{tabular}
\end{table*}

During the data collection process, we controlled the vehicle using an \gls{lmpc} controller and collected two laps of on-track data, resulting in 8,933 data points as training data for \gls{rkmpc}. For \gls{kmpc}, we used randomly generated trajectory data (50000 data points) to enhance its generalization capability.


After determining the residual state equation, the \gls{rkmpc} was combined with the original \gls{mpc} controller. As shown in Fig. \ref{fig.racetrack-database} and Fig. \ref{fig.Comparison of the Three Controllers for One Lap}, the green line represents \gls{rkmpc}, the blue line represents \gls{lmpc}, the red line represents \gls{kmpc}, the cyan line represents \gls{nmpc}. The specific data is shown in Table \ref{table_1}.

The results show that, when \gls{rkmpc} is compared to the \gls{lmpc} (baseline), the lateral error is reduced by 11.21\%, the heading error is reduced by 8.63\%, the front angle rate is reduced by 27.58\%, indicating a notable improvement in both accuracy and stability. As for \gls{nmpc}, although its lateral error is better than others, its maximum computation time reached 66.38 ms, which is four times longger than \gls{rkmpc} and exceeds the 50 ms control cycle time.

\begin{figure}[htbp]
  \centering
  \includegraphics[width = 8cm]{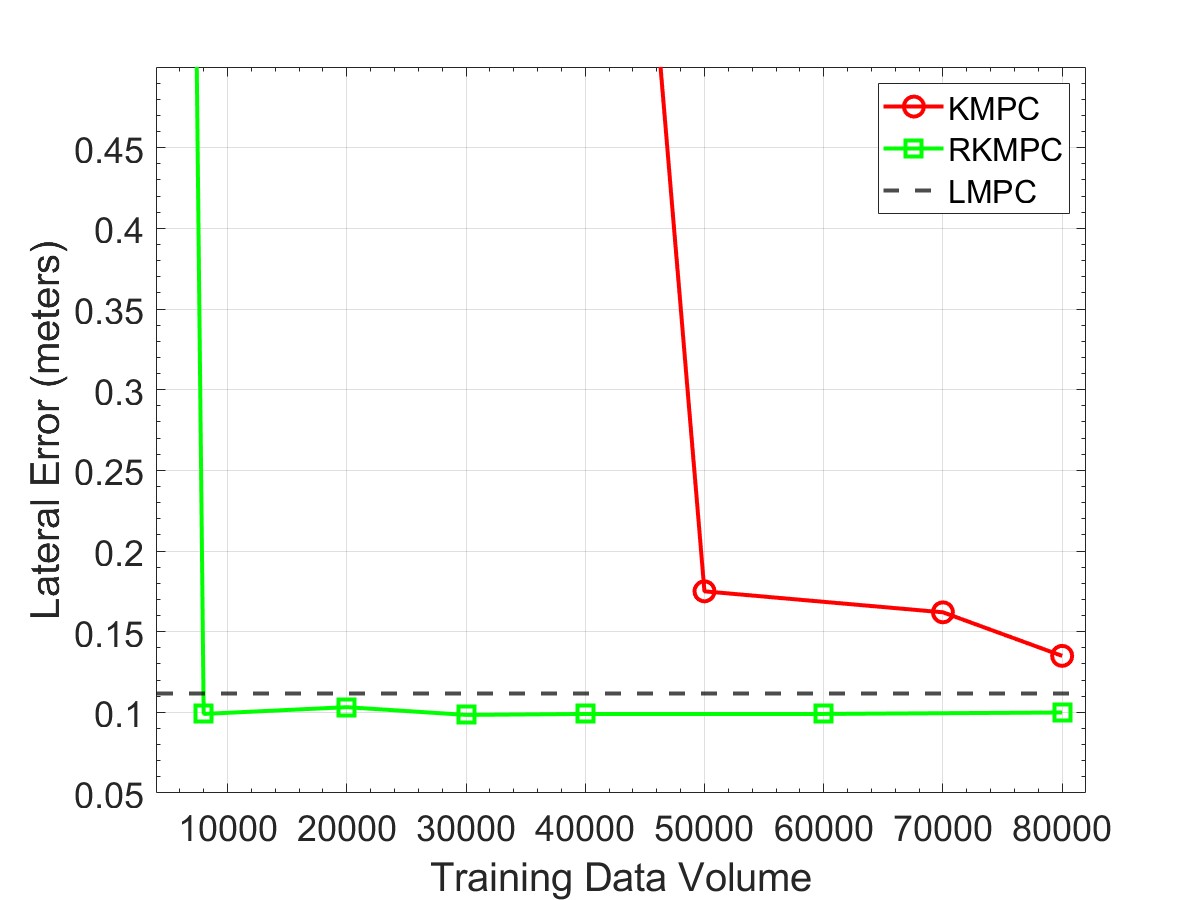}
  \caption{\gls{rkmpc} and \gls{kmpc}'s different requirements for data volume.}
  \label{fig/The Impact of Data Volume on Control Performance}
\end{figure}
As shown in Fig. \ref{fig/The Impact of Data Volume on Control Performance}, \gls{rkmpc} uses about 8000 data points while \gls{kmpc} need 50000 or more data points. \gls{rkmpc} requiring only about 20\% of the training data needed by \gls{kmpc} while maintaining stable control performance, which demonstrates advantages in data efficiency. It effectively reduces the lateral tracking error of the vehicle trajectory, demonstrating better control performance than \gls{kmpc} under small-data conditions.

\section{EXPERIMENTAL RESULTS}

\begin{table*}[htbp]
\caption{Comparison of front wheel angle, lateral error, heading error and computation time among \gls{lmpc}, \gls{kmpc}, and \gls{rkmpc} in real map.}
\label{table_2}
\centering
\renewcommand{\arraystretch}{1.5} 
\begin{threeparttable}
\begin{tabular}{@{}l|cccc@{}}
\toprule
\textbf{Method} & \textbf{Lateral Error (m)} & \textbf{Heading Error (rad)} & \textbf{Front Wheel Angle Rate (rad/s)} & \textbf{Computation Time (ms)} \\ \midrule
\gls{lmpc} (baseline) & 0.284               & 0.1441                  & 0.213                  & \textbf{2.23}       \\ 
\gls{kmpc}\tnote{*}   & -                   & -                       & -                      & 6.54                \\ 
\gls{rkmpc}           & \textbf{0.2213 ($\downarrow$ 22.08\%)} & \textbf{0.1213 ($\downarrow$ 15.82\%)} & \textbf{0.2113 ($\downarrow$ 0.80\%)} & 8.55 \\ \bottomrule
\end{tabular}
\begin{tablenotes}
\footnotesize
\item[*] The symbol `-` indicates that \gls{kmpc} could not complete a full lap.
\end{tablenotes}
\end{threeparttable}
\end{table*}

\begin{figure}[htbp]
  \centering
  \includegraphics[width = 7cm]{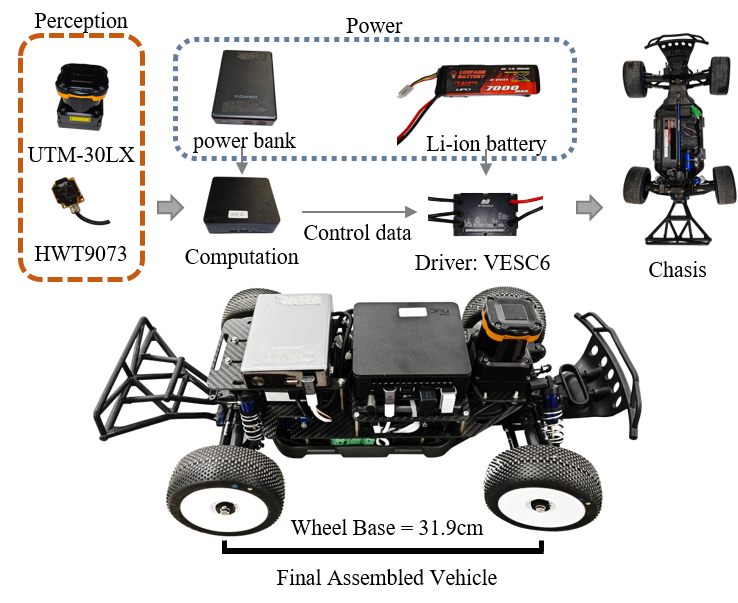}
  \caption{Structural diagram of the F1TENTH vehicle model: components for perception, computation, power, and control.}
  \label{fig.F1TENTH Vehicle Model Structural Diagram}
\end{figure}

\begin{figure}[htbp]
        \centering
        \subcaptionbox{Real Map\label{fig.RealMap}}{
            \includegraphics[width = .45\linewidth]{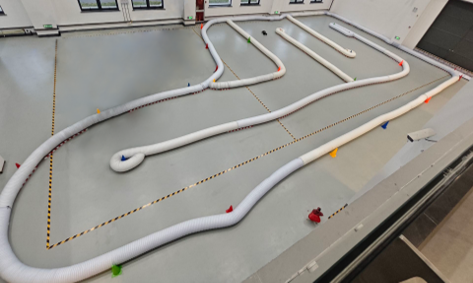}
        }
        \subcaptionbox{Rviz Map\label{fig.RvizMap}}{
            \includegraphics[width = .45\linewidth]{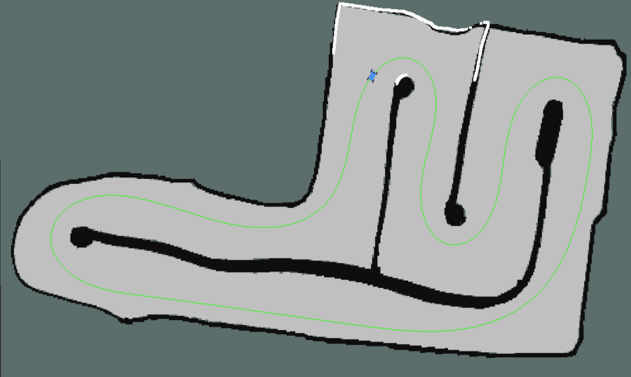}
        }
        \caption{The real map and its visualization in rviz.}
        \label{fig.Real Map and the Map Displayed in RIVZ}
\end{figure}

In this section, we analyze the application performance of the \gls{rkmpc} method on an actual vehicle.
In the experiment, we built a 1:10 scale autonomous vehicle based on the F1TENTH software system\cite{o2020f1tenth}. The computational unit uses an NUC device running ROS Noetic on Ubuntu 20.04, The final vehicle structure is shown in Fig. \ref{fig.F1TENTH Vehicle Model Structural Diagram}. Map construction was performed using IMU, 2D LiDAR, and odometry data through Cartographer\cite{hess2016real}, and localization was achieved using a particle filter\cite{walsh2018cddt}. As shown in the figures, Fig. \ref{fig.Real Map and the Map Displayed in RIVZ} illustrates the racetrack we constructed.


\begin{figure}[htbp]
  \centering
  \includegraphics[width = 7cm]{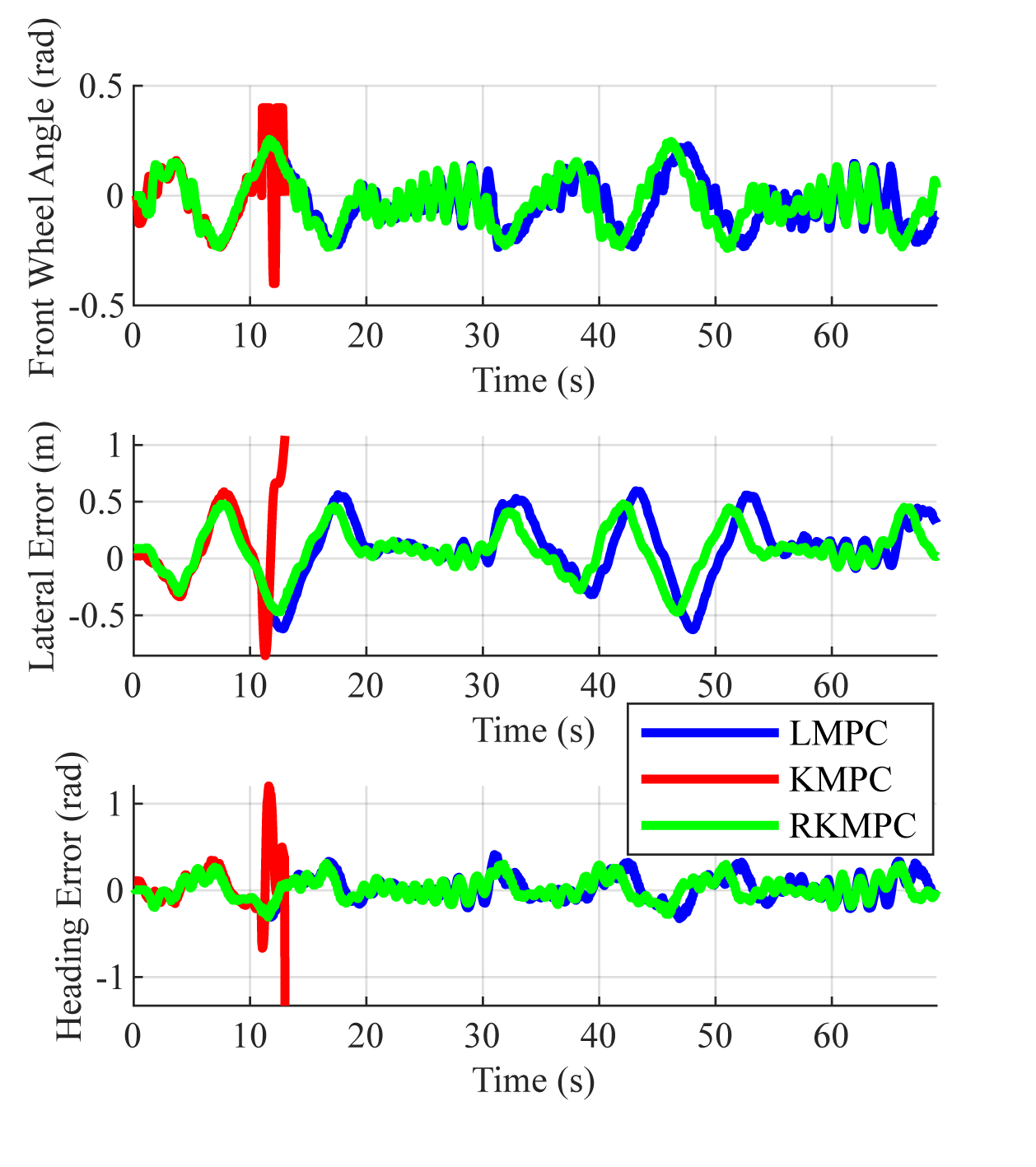}
  \caption{Comparison of front wheel angle, lateral error, and heading error among \gls{lmpc}, \gls{kmpc}, and \gls{rkmpc} in real map. Due the limited scenario data, \gls{kmpc} can not control car stably.}
  \label{fig.Comparison of the Three Controllers in Real Map}
\end{figure}

\begin{figure}[htbp]
  \centering
  \includegraphics[width = 7cm]{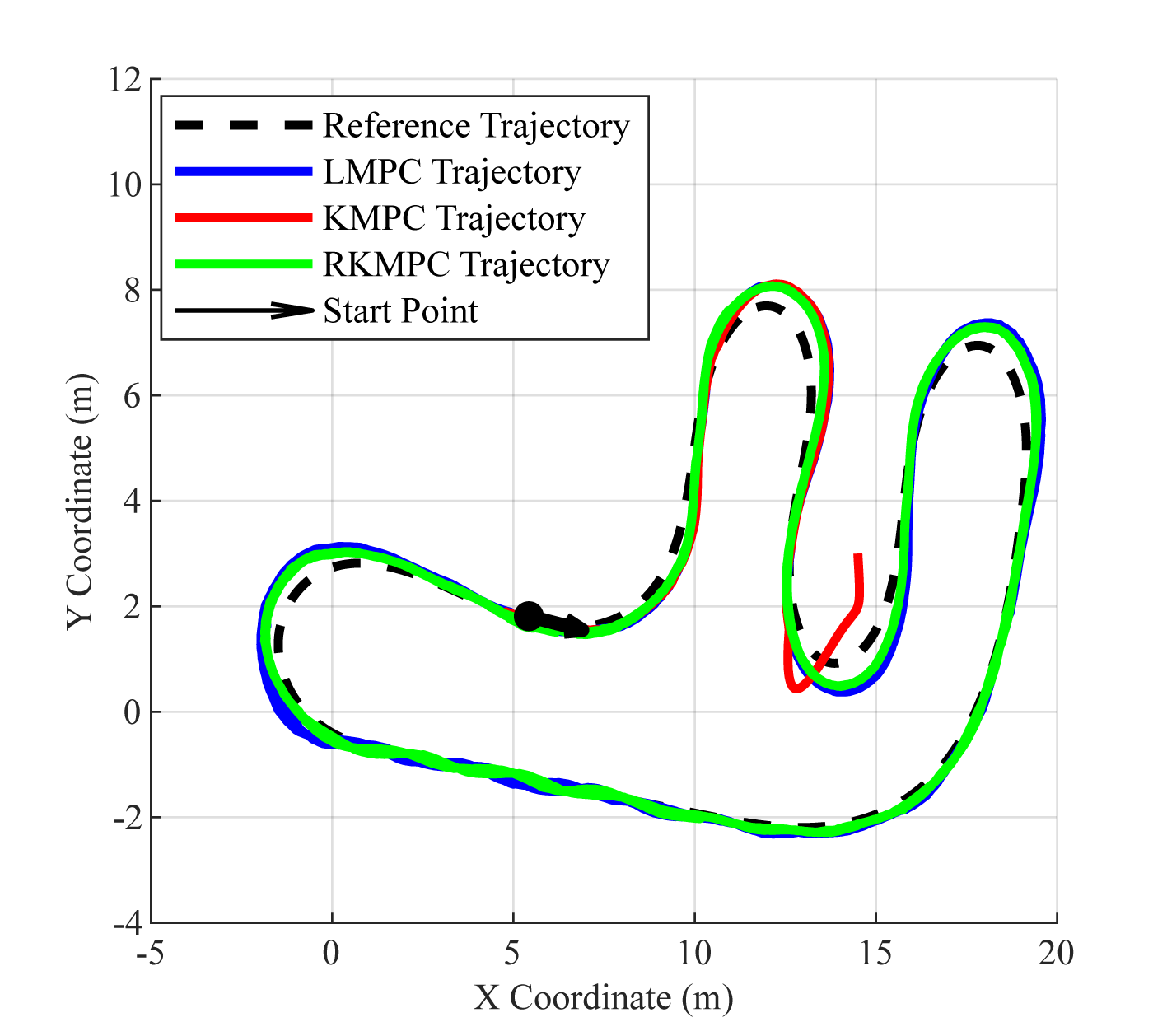}
  \caption{Performance comparison of trajectories controlled by \gls{lmpc}, \gls{kmpc}, and \gls{rkmpc} in real map.}
  \label{fig.Control Mode Comparison in real}
\end{figure}


In physical experiment, we use the same setup to simulation. We collect two laps on-track data in the real map with \gls{lmpc} controller, obtaining 1,527 sets of raw data. By applying a 30\% data conversion ratio, the processed dataset was expanded to 22,950 sets. The final results, as shown in Fig. \ref{fig.Comparison of the Three Controllers in Real Map} and Fig. \ref{fig.Control Mode Comparison in real}, demonstrate that compared to \gls{lmpc}, the \gls{rkmpc}'s vehicle's lateral error was reduced by 22.08\%, heading error decreased by 15.8\%, and stability showed slight improvement. Detailed data can be found in Table \ref{table_2}.

For \gls{kmpc}, we collected data over 10 laps. However, as \gls{kmpc} is a fully data-driven model, relying solely on one type of track configuration proved insufficient for fitting a complete vehicle model, resulting in significant errors in the linear state equations. Once the vehicle's state deviated from the data range, the system became uncontrollable.

Regarding \gls{nmpc}, since its maximum computation time exceeded 50 ms, it couldn't meet real-time control requirements. Therefore, we only conducted simulations for \gls{nmpc} and did not perform physical experiments.

\section{CONCLUSIONS}

This paper proposes a novel \gls{rkmpc} framework that combines a \gls{lmpc} and a neural network-based residual Koopman
\gls{mpc}. The dual-channel control architecture employs both kinematic model-based baseline control and data-driven residual compensation, achieving performance improvements while requiring only 20\% of the training data compared to traditional Koopman methods. Experimental results show that RKMPC reduces lateral error by 11.7\%-22.1\%, heading error by 8.9\%-15.8\%, and improves steering stability by up to 27.6\% compared to \gls{lmpc}. Future research will focus on adapting this method to complex dynamic environments and enhancing real-time control performance through lightweight network architecture design.



\bibliographystyle{unsrt}
\input{root.bbl}

\end{document}

%% file: acr.tex
\newacronym{mpc}{MPC}{Model Predictive Control}
\newacronym{lmpc}{LMPC}{Linear Model Predictive Control}
\newacronym{nmpc}{NMPC}{Nonlinear Model Predictive Control}
\newacronym{kmpc}{KMPC}{Koopman Model Predictive Control}
\newacronym{rkmpc}{RKMPC}{Residual Koopman Model Predictive Control}
\newacronym{edmd}{EDMD}{Extended Dynamic Mode Decomposition}
\newacronym{eso}{ESO}{Extended State Observer}
\newacronym{dnn}{DNN}{Deep Neural Networks}
\newacronym{lqg}{LQG}{Linear Quadratic Gaussian}
\newacronym{pp}{PP}{Pure Pursuit}

%% file: root.bbl

%% file: root.bbl
\begin{thebibliography}{10}
\providecommand{\url}[1]{#1}
\csname url@samestyle\endcsname
\providecommand{\newblock}{\relax}
\providecommand{\bibinfo}[2]{#2}
\providecommand{\BIBentrySTDinterwordspacing}{\spaceskip=0pt\relax}
\providecommand{\BIBentryALTinterwordstretchfactor}{4}
\providecommand{\BIBentryALTinterwordspacing}{\spaceskip=\fontdimen2\font plus
\BIBentryALTinterwordstretchfactor\fontdimen3\font minus \fontdimen4\font\relax}
\providecommand{\BIBforeignlanguage}[2]{{%
\expandafter\ifx\csname l@#1\endcsname\relax
\typeout{** WARNING: IEEEtran.bst: No hyphenation pattern has been}%
\typeout{** loaded for the language `#1'. Using the pattern for}%
\typeout{** the default language instead.}%
\else
\language=\csname l@#1\endcsname
\fi
#2}}
\providecommand{\BIBdecl}{\relax}
\BIBdecl


\bibitem{8315037}
H.~Guo, C.~Shen, H.~Zhang, H.~Chen, and R.~Jia, ``Simultaneous trajectory planning and tracking using an mpc method for cyber-physical systems: A case study of obstacle avoidance for an intelligent vehicle,'' \emph{IEEE Transactions on Industrial Informatics}, vol.~14, no.~9, pp. 4273--4283, 2018.

\bibitem{kong2015kinematic}
J.~Kong, M.~Pfeiffer, G.~Schildbach, and F.~Borrelli, ``Kinematic and dynamic vehicle models for autonomous driving control design,'' in \emph{2015 IEEE intelligent vehicles symposium (IV)}.\hskip 1em plus 0.5em minus 0.4em\relax IEEE, 2015, pp. 1094--1099.

\bibitem{korda2018linear}
M.~Korda and I.~Mezi{\'c}, ``Linear predictors for nonlinear dynamical systems: Koopman operator meets model predictive control,'' \emph{Automatica}, vol.~93, pp. 149--160, 2018.

\bibitem{cibulka2020model}
V.~Cibulka, T.~Hani{\v{s}}, M.~Korda, and M.~Hrom{\v{c}}{\'\i}k, ``Model predictive control of a vehicle using koopman operator,'' \emph{IFAC-PapersOnLine}, vol.~53, no.~2, pp. 4228--4233, 2020.

\bibitem{sinha2020robust}
S.~Sinha, B.~Huang, and U.~Vaidya, ``On robust computation of koopman operator and prediction in random dynamical systems,'' \emph{Journal of Nonlinear Science}, vol.~30, no.~5, pp. 2057--2090, 2020.

\bibitem{hoffmann2007autonomous}
G.~M. Hoffmann, C.~J. Tomlin, M.~Montemerlo, and S.~Thrun, ``Autonomous automobile trajectory tracking for off-road driving: Controller design, experimental validation and racing,'' in \emph{2007 American control conference}.\hskip 1em plus 0.5em minus 0.4em\relax IEEE, 2007, pp. 2296--2301.

\bibitem{wang2017improved}
W.-J. Wang, T.-M. Hsu, and T.-S. Wu, ``The improved pure pursuit algorithm for autonomous driving advanced system,'' in \emph{2017 IEEE 10th international workshop on computational intelligence and applications (IWCIA)}.\hskip 1em plus 0.5em minus 0.4em\relax IEEE, 2017, pp. 33--38.

\bibitem{li2024data}
Z.~Li, B.~Zhou, C.~Hu, L.~Xie, and H.~Su, ``A data-driven aggressive autonomous racing framework utilizing local trajectory planning with velocity prediction,'' \emph{arXiv preprint arXiv:2410.11570}, 2024.

\bibitem{hu2022combined}
C.~Hu, X.~Zhou, R.~Duo, H.~Xiong, Y.~Qi, Z.~Zhang, and L.~Xie, ``Combined fast control of drifting state and trajectory tracking for autonomous vehicles based on mpc controller,'' in \emph{2022 International Conference on Robotics and Automation (ICRA)}.\hskip 1em plus 0.5em minus 0.4em\relax IEEE, 2022, pp. 1373--1379.

\bibitem{bienemann2023model}
A.~Bienemann and H.-J. Wuensche, ``Model predictive control for autonomous vehicle following,'' in \emph{2023 IEEE Intelligent Vehicles Symposium (IV)}.\hskip 1em plus 0.5em minus 0.4em\relax IEEE, 2023, pp. 1--6.

\bibitem{gao2014tube}
Y.~Gao, A.~Gray, H.~E. Tseng, and F.~Borrelli, ``A tube-based robust nonlinear predictive control approach to semiautonomous ground vehicles,'' \emph{Vehicle System Dynamics}, vol.~52, no.~6, pp. 802--823, 2014.

\bibitem{zhang2022residual}
R.~Zhang, J.~Hou, G.~Chen, Z.~Li, J.~Chen, and A.~Knoll, ``Residual policy learning facilitates efficient model-free autonomous racing,'' \emph{IEEE Robotics and Automation Letters}, vol.~7, no.~4, pp. 11\,625--11\,632, 2022.

\bibitem{trumpp2023residual}
R.~Trumpp, D.~Hoornaert, and M.~Caccamo, ``Residual policy learning for vehicle control of autonomous racing cars,'' in \emph{2023 IEEE Intelligent Vehicles Symposium (IV)}.\hskip 1em plus 0.5em minus 0.4em\relax IEEE, 2023, pp. 1--6.

\bibitem{long2025physical}
K.~Long, Z.~Sheng, H.~Shi, X.~Li, S.~Chen, and S.~Ahn, ``Physical enhanced residual learning (perl) framework for vehicle trajectory prediction,'' \emph{Communications in Transportation Research}, vol.~5, p. 100166, 2025.

\bibitem{brunton2016koopman}
S.~L. Brunton, B.~W. Brunton, J.~L. Proctor, and J.~N. Kutz, ``Koopman invariant subspaces and finite linear representations of nonlinear dynamical systems for control,'' \emph{PloS one}, vol.~11, no.~2, p. e0150171, 2016.

\bibitem{xiao2022deep}
Y.~Xiao, X.~Zhang, X.~Xu, X.~Liu, and J.~Liu, ``Deep neural networks with koopman operators for modeling and control of autonomous vehicles,'' \emph{IEEE transactions on intelligent vehicles}, vol.~8, no.~1, pp. 135--146, 2022.

\bibitem{yakub2015comparative}
F.~Yakub and Y.~Mori, ``Comparative study of autonomous path-following vehicle control via model predictive control and linear quadratic control,'' \emph{Proceedings of the Institution of Mechanical Engineers, Part D: Journal of automobile engineering}, vol. 229, no.~12, pp. 1695--1714, 2015.

\bibitem{wu2024learning}
G.~Wu, C.~Hu, W.~Weng, Z.~Li, Y.~Fu, L.~Xie, and H.~Su, ``Learning to race in extreme turning scene with active exploration and gaussian process regression-based mpc,'' \emph{arXiv preprint arXiv:2410.05740}, 2024.

\bibitem{mauroy2020introduction}
A.~Mauroy, Y.~Susuki, and I.~Mezi{\'c}, ``Introduction to the koopman operator in dynamical systems and control theory,'' \emph{The koopman operator in systems and control: concepts, methodologies, and applications}, pp. 3--33, 2020.

\bibitem{wang2021deep}
R.~Wang, Y.~Han, and U.~Vaidya, ``Deep koopman data-driven control framework for autonomous racing,'' in \emph{Proc. Int. Conf. Robot. Autom.(ICRA) Workshop Opportunities Challenges Auton. Racing}, 2021, pp. 1--6.

\bibitem{racetrack-database}
T.~I. of~Automotive~Technology, ``racetrack-database,'' \url{https://github.com/TUMFTM/racetrack-database}, 2021.

\bibitem{o2020f1tenth}
M.~O'Kelly, H.~Zheng, D.~Karthik, and R.~Mangharam, ``F1tenth: An open-source evaluation environment for continuous control and reinforcement learning,'' \emph{Proceedings of Machine Learning Research}, vol. 123, 2020.

\bibitem{hess2016real}
W.~Hess, D.~Kohler, H.~Rapp, and D.~Andor, ``Real-time loop closure in 2d lidar slam,'' in \emph{2016 IEEE international conference on robotics and automation (ICRA)}.\hskip 1em plus 0.5em minus 0.4em\relax IEEE, 2016, pp. 1271--1278.

\bibitem{walsh2018cddt}
C.~H. Walsh and S.~Karaman, ``Cddt: Fast approximate 2d ray casting for accelerated localization,'' in \emph{2018 IEEE International Conference on Robotics and Automation (ICRA)}.\hskip 1em plus 0.5em minus 0.4em\relax IEEE, 2018, pp. 3677--3684.

































\end{thebibliography}
